\documentclass[11pt]{article}

\usepackage[final]{acl}

\usepackage{times}
\usepackage{latexsym}

\usepackage[T1]{fontenc}
\usepackage[utf8]{inputenc}

\usepackage{microtype}

\usepackage{inconsolata}

\usepackage{graphicx}
\usepackage{booktabs}
\usepackage{multirow}
\usepackage{amsmath}
\usepackage{tcolorbox}

\title{SCURank: Ranking Multiple Candidate Summaries with Summary Content Units for Enhanced Summarization}

\author{
  \textbf{Bo-Jyun Wang\textsuperscript{1,2}},
  \textbf{Ying-Jia Lin\textsuperscript{2,3}},
  \textbf{Hung-Yu Kao\textsuperscript{4}}\\
  \textsuperscript{1}Department of Computer Science and Information Engineering, \\National Cheng Kung University,\\
  \textsuperscript{2}Artificial Intelligence Research Center, Chang Gung University,\\
  \textsuperscript{3}Department of Artificial Intelligence, Chang Gung University,\\
  \textsuperscript{4}Department of Computer Science, National Tsing Hua University\\
  \texttt{bojyun.wang@cgu.edu.tw}, \texttt{yjlin@cgu.edu.tw}, \texttt{hykao@cs.nthu.edu.tw}
}

\begin{document}
\maketitle
\begin{abstract}
  Small language models (SLMs), such as BART, can achieve summarization performance comparable to large language models (LLMs) via distillation.
  However, existing LLM-based ranking strategies for summary candidates suffer from instability, while classical metrics (e.g., ROUGE) are insufficient to rank high-quality summaries.
  To address these issues, we introduce \textbf{SCURank}, a framework that enhances summarization by leveraging \textbf{Summary Content Units (SCUs)}.
  Instead of relying on unstable comparisons or surface-level overlap, SCURank evaluates summaries based on the richness and semantic importance of information content.
  We investigate the effectiveness of SCURank in distilling summaries from multiple diverse LLMs.
  Experimental results demonstrate that SCURank outperforms traditional metrics and LLM-based ranking methods across evaluation measures and datasets.
  Furthermore, our findings show that incorporating diverse LLM summaries enhances model abstractiveness and overall distilled model performance, validating the benefits of information-centric ranking in multi-LLM distillation.
  The code for SCURank is available at \url{https://github.com/IKMLab/SCURank}.
\end{abstract}

\section{Introduction}
Following the advent of ChatGPT \citep{ouyang2022training}, a paradigm shift has occurred in the domain of Natural Language Processing (NLP), marked by substantial advancements, including the field of summarization.
In the wake of this development, GPT-4 \citep{openai2024gpt4technicalreport} and numerous Large Language Models (LLMs) have emerged \citep{gemini_models, claude2024, mistral2024}, demonstrating superior performance.
These models are easily accessible via APIs, allowing users to obtain responses with minimal effort.
However, while these models are powerful, their resource demands, such as the cost of local deployments, are significant \citep{hsieh-etal-2023-distilling} due to their incredibly large model size.
This has led to a growing trend of distilling these models into smaller ones, optimized for specific tasks such as summarization \cite{hinton2015distillingknowledgeneuralnetwork,jiang-etal-2024-trisum}.
The distilled models exhibit great resource efficiency without significant performance degradation, and in some cases, they demonstrate a capability to outperform LLMs in the summarization task \cite{liu-etal-2024-learning}.

Previous work by \citet{liu-etal-2024-learning} leverages BRIO \citep{liu-etal-2022-brio} as a contrastive learning framework for model distillation.
In BRIO, positive and negative samples are constructed by ranking candidate summaries.
Consequently, the quality of the ranking function is critical for effective contrastive learning.
% They used summaries generated by GPT-3.5 and GPT-4 as references, along with candidate summaries generated by an LLM through diverse beam search \citep{vijayakumar2016diverse}.
% Since BRIO \citep{liu-etal-2022-brio} relies on contrastive learning, effectively ranking the candidate summaries is crucial for the training process.
To address this challenge, they introduced GPTRank \cite{liu-etal-2024-learning}, a novel method that ranks high-quality summaries generated by LLMs, ensuring more reliable supervision in BRIO training.
However, there are two significant challenges to this approach.
First, studies by \citet{shen-etal-2023-large, wang-etal-2024-large-language-models-fair} indicate that LLMs remain unreliable and inconsistent in text comparison and candidate ranking.
Second, relying on summaries from a single LLM introduces the risk of model-specific bias (e.g. content selection) and limits the diversity of generation patterns.
Thus, we explore the use of multiple LLMs to generate candidate summaries for distillation, and investigate effective ranking methods for these summaries.

To bypass the instability of direct LLM ranking, we propose shifting the evaluation focus back to the core goal of summarization: information retention.
To capture the information from summaries, we draw upon the concept of Summary Content Units (SCUs) \cite{nenkova-passonneau-2004-evaluating}.
Every SCU presents simple, standalone, and unique information in the summary \citep{shapira-etal-2019-crowdsourcing}.
Generally, annotating SCUs requires manual efforts, thus expensive and not reproducible \citep{zhang-bansal-2021-finding}.
\citet{nawrath-etal-2024-role} suggested a new path to extract SCUs, which employed GPT-3.5 and GPT-4 \citep{openai2024gpt4technicalreport} to generate Semantic GPT Units (SGUs).
The SGUs are similar to SCUs, capturing the key information in the summaries, and have been proven to be high-quality in the evaluations of \citet{nawrath-etal-2024-role}.
Consequently, we adopt the concept of SGUs as SCUs to evaluate the quality of summaries.

In this paper, we introduce \textbf{SCURank} (\textbf{S}ummary \textbf{C}ontent \textbf{U}nit \textbf{Rank}ing), a ranking framework that evaluates the information richness of candidate summaries by analyzing their SCUs.
SCURank operates in three stages.
First, it captures the key information in each summary by extracting SCUs.
Next, all SCUs are aggregated by clustering to estimate their importance based on their frequency across summaries.
Finally, each summary is assigned a score by summing the importance scores of its SCUs, which reflects its overall information richness.
This score is further normalized by summary length to mitigate bias toward longer summaries.
By focusing on the information content rather than direct comparison or ranking, SCURank provides a robust ranking approach.
Specifically, as LLMs are only employed for SCU extraction, the framework avoids the unreliability associated with LLM-based comparison.

Our contributions are as follows:
(1) We propose SCURank, a novel method to rank high-quality summaries based on SCUs.
(2) We investigate the effect of distilling models from multiple LLMs.
(3) We demonstrate that SCURank, combined with contrastive learning, improves distilled model performance.

\begin{figure*}[ht]
  \centering
  \includegraphics[width=\linewidth]{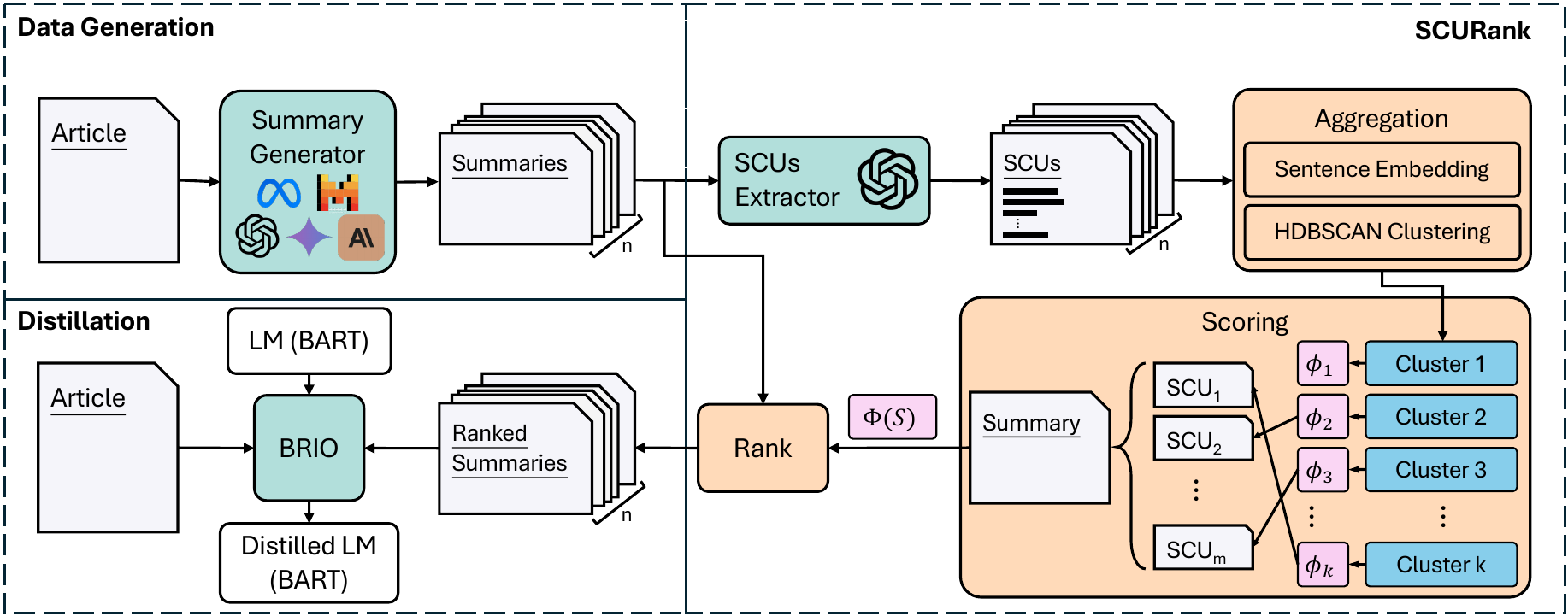}
  \caption{
    Overview of our training framework.
    The Data Generation part generates candidate summaries from several LLMs for each article.
    SCURank, our proposed ranking method, ranks the candidate summaries via three steps: (1) extracting Summary Content Units (SCUs), (2) aggregating SCUs with sentence embeddings and HDBSCAN, and (3) scoring summaries based on SCU cluster distribution.
    Finally, the ranked summaries are then used to train a distilled model with BRIO.
    }
  \label{fig:big-picture}
\end{figure*}

\section{Related Work}
SCURank, which ranks summaries by employing Summary Content Units (SCUs), is based on two concepts.
The first method is to decompose a summary candidate into SCUs, which are brief and convey a single fact.
The second is the ranking.
The ranking results for the candidate summaries can be used for contrastive learning to improve the LMs' summarization capabilities.

\subsection{Summary Content Units}
The concept of SCUs was first introduced by \citet{nenkova-passonneau-2004-evaluating}.
In that study, the authors introduced a reliable, predictive, and diagnostic method for evaluating the summaries.
However, due to the high cost and expertise required, \citet{shapira-etal-2019-crowdsourcing} demonstrated a revised version that is more cost-effective and annotation-friendly.
In this approach, they provided the instructions for crowd workers to extract a specific number of the SCU-like statements, eliminating the need to merge and weight SCUs.
These efforts reduced the dependency on expertise, making the Pyramid method more efficient.

Nevertheless, the significant cost of hiring workers to extract SCUs remains a significant challenge.
In response, researchers have been investigating methods of automating this process in recent years.
\citet{zhang-bansal-2021-finding} proposed a system called $List^3Pyramid$, which employed a semantic role labeling model to extract Summary Triplets Units (STUs).
Subsequently, \citet{nawrath-etal-2024-role} introduced two novel methods.
One method leveraged Abstract Meaning Representation to extract Semantic Meaning Units (SMUs), and the other employed an LLM to extract Semantic GPT Units (SGUs).
\citet{nawrath-etal-2024-role} have demonstrated the high quality of the SGUs in a wide range of evaluations.
In our work, we adopt the concept of SGUs from \citet{nawrath-etal-2024-role} for extracting SCUs in SCURank.

\subsection{Summarization with Ranking}
Contrastive learning is a method of training a model to identify and select superior candidates, while avoiding inferior ones.
In summarization, SimCLS \citep{liu-liu-2021-simcls} was the first to employ a scoring model for contrastive learning.
This scoring model evaluates the quality of the candidate summaries, thereby enhancing overall performance on summarization tasks.
Subsequently, BRIO \citep{liu-etal-2022-brio} integrated these tasks into a single model, which is capable of both generating and evaluating summaries.
Both approaches used ROUGE \citep{lin-2004-rouge} as a metric for evaluating summaries.
However, as ROUGE is only concerned with n-gram overlap, it is not accurate enough to evaluate high-quality summaries \citep{cohan2016revisiting}.
Recently, \citet{liu-etal-2024-learning} developed GPTRank, which integrates the concept of G-Eval \citep{liu-etal-2023-g}, to rank the summaries using an LLM.
In contrast to GPTScore \citep{fu-etal-2024-gptscore}, GPTRank does not employ the LLM-predicted probability.
Instead, it provides a prompt to request the LLM to rank and offer a concise explanation of the ranking.
Nevertheless, \citet{wang-etal-2024-large-language-models-fair} uncovered positional bias in the ranking of LLMs, and \citet{shen-etal-2023-large} also demonstrated that LLMs' difficulty in providing consistent text comparisons.
These studies demonstrate the need for a more concrete and stable method for training summarization models with ranked candidates \cite{liu-etal-2024-learning}.

\section{Methods}
\subsection{Preliminary}
Let $\mathcal{D}$ denote the document being summarized.
Assume that there are $n$ summarization models $F = \{f_i\}_{i=1}^n$ we want to distill.
All models use the same prompt (Appendix~\ref{sec:llm-gen-details}). % Appendix
The set of all summaries produced by these models is given by $S=\{s_i\}_{i=1}^n$, where each summary $s_i$ is generated by the corresponding model $f_i$:
\begin{equation}
  \label{eq:SummaryGeneration}
  s_i = f_i(\mathcal{D}).
\end{equation}
\subsection{SCURank}
\subsubsection{Overview}
The SCURank is designed to rank high-quality summaries based on their information richness and importance.
As illustrated in Figure~\ref{fig:big-picture}, the SCURank process has three key steps: (1) extract SCUs from the candidate summaries, (2) aggregate SCUs across multiple summaries via clustering, and (3) score each summary based on the importance of its SCUs.
The details are below.
\subsubsection{SCUs Extraction}
To extract the SCUs from summaries, we adopt the method based on a large language model from \citet{nawrath-etal-2024-role}.
The instruction contains one example from REALSumm \cite{bhandari-etal-2020-evaluating} to generate SCUs.
We use gpt-4o-mini\footnote{We've compared the performance of gpt-4o-mini, gpt-4o, and gpt-3.5-turbo and found that gpt-4o-mini is cost-effective while maintaining comparable performance to gpt-4o. Details and the preliminary results are provided in the Appendix~\ref{sec:scu-details}.}\cite{gpt4omini} as $\mathrm{SCUExt}$ to extract SCUs from the summaries:
\begin{equation}
  \label{eq:SCUExtract}
  \mathcal{U}_i = \{u_{i,1}, u_{i,2},...,u_{i,m_i}\} = \mathrm{SCUExt}(s_i).
\end{equation}
Each $\mathcal{U}_i$ contains all the SCUs extracted from $s_i$.
The number of SCUs in $s_i$, denoted as $m_i$, varies since the amount of information present in each summary $s_i$ is uncertain.

\subsubsection{SCUs Aggregation}
After extracting SCUs from the candidate summaries, we aggregate all SCUs by clustering them based on their semantic similarity.
Each cluster represents distinct semantic information, and the number of SCUs in a cluster reflects the importance of that information.
The more SCUs in a cluster, the more models agree on the information, thus presenting the level of its value.

\paragraph{Sentence Encoder}
To facilitate clustering, it is essential to translate these SCUs into vectors.
We use the all-mpnet-base-v2\footnote{https://huggingface.co/sentence-transformers/all-mpnet-base-v2} model, noted as $Encoder$, which is small but effective \citep{reimers-2019-sentence-bert}.
Given a set of SCUs $\mathcal{U}_i$, the corresponding vectors are obtained as follows:
\begin{equation}
  \label{eq:SCUEmbedding}
  \mathcal{V}_i = \{v_{i,1}, v_{i,2},...,v_{i,m_i}\} = \mathrm{Encoder}(\mathcal{U}_i).
\end{equation}
$\mathcal{V}_i$ represents the set of vectorized SCUs for $s_i$.
These embeddings capture semantic meaning, allowing us to cluster the SCUs.
We aggregate all embeddings as $\mathcal{V} = \bigcup_{i=1}^n \mathcal{V}_i$ for clustering.

\paragraph{HDBSCAN}
Since the number of distinct semantic information is uncertain, we employ HDBSCAN \citep{10.1007/978-3-642-37456-2_14} for clustering the SCU embeddings.
HDBSCAN (Hierarchical Density-Based Spatial Clustering of Applications with Noise), an extension of DBSCAN \citep{ester1996density}, can automatically determine the number of clusters based on the density of the data and identifies outliers as noise \citep{10.1007/978-3-642-37456-2_14}.
Unlike DBSCAN, HDBSCAN can handle varying cluster densities, making it more suitable for our task.

Our task is to cluster the SCUs into different groups, with each group representing distinct semantic information.
The result of the clustering is a set of clusters, denoted as $\mathcal{C}$.
\begin{equation}
  \label{eq:SCUCluster}
  \mathcal{C} = \mathrm{HDBSCAN}(\mathcal{V})
\end{equation}
\begin{equation}
  \label{eq:SCUCluster-Result}
  \mathcal{C} = \{C^1, C^2,..., C^K\}
\end{equation}
Each $C^k (k\in {1, \dots, K})$ contains the SCUs that belong to the same cluster.

\subsubsection{Scoring}
Our scoring mechanism is based on the assumption that information selected by more candidate summaries is more important.
Since SCUs within the same cluster $C^k$ are semantically equivalent, the cluster size reflects the level of consensus among candidates.
We thus define the score of each SCU $u_{i,j}$ as:
\begin{equation}
  \label{eq:SCUScore}
  \phi(u_{i,j}) = \|C^k\|, \text{ where } u_{i,j} \in C^k.
\end{equation}
For example, if $u_{i,j}$ belongs to a cluster $C^k$ that contains 5 SCUs, then $\phi(u_{i,j}) = 5$.

For each summary candidate $s_i$, the function $\Phi$ returns the sum of the scores of the information associated with its SCUs:
\begin{equation}
  \label{eq:SummaryScore}
  \Phi(s_i) = \sum_{j=1}^{m_i} \phi(u_{i,j}).
\end{equation}
To prevent longer summaries from receiving disproportionately higher scores, we divide all scores by the square root of the number of tokens in each summary.
This produces the final adjusted scores with length penalty (lp):
\begin{equation}
  \label{eq:LengthPenalty}
  \Phi_\mathrm{lp}(s_i) = \frac{\Phi(s_i)}{\sqrt{\|s_i\|}},
\end{equation}
where $\|s_i\|$ is the number of tokens in the summary $s_i$.
Preliminary results are provided in Appendix~\ref{sec:length_penalty_analysis}. % Appendix

\subsubsection{Rank}
With the adjusted scores, we can rank the summaries based on the richness and importance of their SCUs.
The standard ranking is defined as follows:
\begin{equation}
  \label{eq:SummaryRank}
  \mathrm{rk} = \mathrm{argsort}(\Phi_\mathrm{lp}(s_i) \text{ for } i \in \{1, 2, ..., n\})
\end{equation}
so that:
\begin{equation}
  \label{eq:SummaryRank-Result}
  \resizebox{0.86\hsize}{!}{
    $\Phi_\mathrm{lp}(S_{\mathrm{rk}(1)}) \ge \Phi_\mathrm{lp}(S_{\mathrm{rk}(2)}) \geq ... \geq \Phi_\mathrm{lp}(S_{\mathrm{rk}(n)}).$
  }
\end{equation}
With this standard ranking result, we can rank the summaries and use them in subsequent training.

\subsection{Distillation}
We integrate SCURank into the BRIO \citep{liu-etal-2022-brio} framework to enhance contrastive learning-based distillation.
BRIO requires a ranking method to differentiate summary quality, and we replace its original ROUGE-based ranking with SCURank to enable more effective training.

\subsubsection{Contrastive Learning}
\label{sec:contrastive learning}
BRIO leverages multiple candidate summaries via contrastive learning to help the model distinguish between high-quality and low-quality summaries.
Given two summaries $s_1$, $s_2$, the model should assign a higher probability to $s_1$ if it is of higher quality than $s_2$.
Initially, BRIO employed an automatic metric such as ROUGE \citep{lin-2004-rouge} to rank the summaries.
\citet{liu-etal-2024-learning} later introduced GPTRank, which replaced ROUGE with GPT-based ranking, but it suffers from instability and inconsistency.
To address these issues, we substitute BRIO's ranking method with SCURank, which offers a more stable and semantically meaningful ranking mechanism.

\subsubsection{Maximum Likelihood Estimation}
We also train a distilled model using maximum likelihood estimation (MLE) to compare its effectiveness with contrastive learning.
Within the BRIO training framework, contrastive learning requires ranked candidate summaries, whereas MLE does not.
Instead, MLE optimizes the likelihood of the reference summaries $s^*$ by minimizing the cross-entropy loss:
\begin{equation}
  \label{eq:MLE}
  \mathcal{L}_{\mathrm{xent}}(\theta) = -\log (\prod_{l=1}^{|s^*|} p(s^*_l | s^*_{<l}, \mathcal{D}; \theta)).
\end{equation}
Here $s^*_l$ denotes the $l$th token in the summary $s^*$, and $p$ is the probability of $s^*_l$ given the previous tokens $s^*_{<l}$ and the document $\mathcal{D}$.
The learnable parameters of the model are represented by the symbol $\theta$.

\section{Experiments}
We conduct several experiments to evaluate the effectiveness and characteristics of our approach.
(1) Distilled model evaluations: We assess the performance of distilled models trained using different ranking methods and automatic metrics.
(2) Stability of SCURank: To analyze the ranking consistency of SCURank and GPTRank, we rank the training set multiple times and measured the correlation.
(3) LLM-based comparison: We use recent state-of-the-art LLMs to compare summaries generated by the distilled models.
(4) Human evaluation: The performance of SCURank is compared against GPTRank using MTurk for pairwise comparison in three dimensions.
(5) Writing style: We examine the abstractiveness of summaries generated by the distilled models and investigate the impact of different training datasets.
Implementation details are provided in the Appendix~\ref{sec:implementation-details}. % Appendix

\subsection{Training Set}
We use two datasets to compare distillation from a single LLM versus multiple LLMs:
The first dataset, \textbf{BASE}, contains summaries generated by GPT-3.5-turbo and candidate summaries from a single, unspecified LLM \cite{liu-etal-2024-learning}.
The second dataset, \textbf{LLMs-9}, consists of summaries generated by nine different LLMs.
The statistical details of both datasets are presented in Appendix~\ref{sec:data-stat}. % Appendix

\subsubsection{BASE}
\label{sec:BASE-dataset}
The BASE dataset was derived from \citet{liu-etal-2024-learning}, which contains 1,000 articles from CNN/DailyMail \citep{nallapati2016abstractive}.
Each article includes a reference generated from gpt-3.5-turbo and nine candidate summaries produced by a single LLM via diverse beam search \citep{vijayakumar2016diverse}.
However, \citet{liu-etal-2024-learning} did not specify the exact model used to generate the summaries.
Nevertheless, we consider BASE to be a single-LLM dataset because the summaries are generated by a single, albeit unspecified, LLM.

\subsubsection{LLMs-9}
\label{sec:LLMs-generated-Summaries}
For a fair comparison, we generated summaries for the same 1,000 CNN/DailyMail articles as in the BASE dataset, using \textbf{nine different LLMs}.
We refer to this collection as \textbf{LLMs-9}.
Details of the adopted models and prompting strategies are provided in the Appendix~\ref{sec:llm-gen-details}.

To further evaluate the effectiveness of SCURank, we extended our experiments to a different summarization dataset, XSum \citep{narayan-etal-2018-dont}.
Using the same nine LLMs, we generated summaries for 1,000 articles from the XSum validation set.
Note that \citet{liu-etal-2024-learning} did not include XSum in the original BASE dataset.

\subsection{Baselines}
We compare SCURank against several baselines and metrics: \textbf{LLMs Average}, \textbf{UnRank} (MLE-only), \textbf{GPTRank} \citep{liu-etal-2024-learning}, \textbf{ROUGE} \citep{lin-2004-rouge}, \textbf{BERTScore} \citep{zhang2019bertscore}, and \textbf{BLANC} \citep{vasilyev-etal-2020-fill}.
For GPTRank, we used gpt-4o-mini to ensure a fair comparison with our method.

\begin{table*}[!t]
  \centering
  \resizebox{0.92\textwidth}{!}{
  \begin{tabular}{l|cccccc}
    \toprule
                     & ROUGE-1                                   & ROUGE-2                                   & ROUGE-L                                   & BLEURT                                    & BERTScore                                 & BARTScore                                  \\
    \midrule\midrule
    \multicolumn{7}{c}{LLMs Results}                                                                                                                                                                                                                                                          \\
    \midrule
    GPT-4o           & 41.2                                      & 15.9                                      & 26.0                                      & 54.7                                      & 69.4                                      & -2.78                                      \\
    Gemini-1.5-pro   & 41.5                                      & 16.7                                      & 27.2                                      & 53.9                                      & 69.7                                      & -2.67                                      \\
    Mistral-Large    & 45.1                                      & 18.8                                      & 29.4                                      & 55.9                                      & 71.2                                      & -2.48                                      \\
    LLMs Average     & 42.0                                      & 17.3                                      & 27.2                                      & 54.5                                      & 69.7                                      & -2.59                                      \\
    \midrule\midrule
    \multicolumn{7}{c}{Dataset: LLMs-9}                                                                                                                                                                                                                                                       \\
    \midrule
    UnRank           & 43.8\textsubscript{$\pm$0.83}             & \textbf{20.5}\textsubscript{$\pm$0.46}    & 30.1\textsubscript{$\pm$0.93}             & \textbf{53.3}\textsubscript{$\pm$0.27}    & \underline{69.9}\textsubscript{$\pm$0.27} & -2.59\textsubscript{$\pm$0.47}             \\
    \midrule
    ROUGE            & 43.2\textsubscript{$\pm$0.23}             & \underline{20.4}\textsubscript{$\pm$0.57} & 29.9\textsubscript{$\pm$0.22}             & 51.4\textsubscript{$\pm$0.40}             & 69.0\textsubscript{$\pm$0.05}             & -2.58\textsubscript{$\pm$0.02}             \\
    BERTScore        & 43.9\textsubscript{$\pm$0.59}             & 20.0\textsubscript{$\pm$0.58}             & 30.1\textsubscript{$\pm$0.81}             & \underline{52.2}\textsubscript{$\pm$0.50} & 69.5\textsubscript{$\pm$0.32}             & -2.46\textsubscript{$\pm$0.06}             \\
    BLANC            & \underline{44.1}\textsubscript{$\pm$0.54} & 20.3\textsubscript{$\pm$0.47}             & \underline{30.2}\textsubscript{$\pm$0.35} & 50.1\textsubscript{$\pm$0.64}             & 69.4\textsubscript{$\pm$0.52}             & \textbf{-2.34}\textsubscript{$\pm$0.13}    \\
    GPTRank          & 43.8\textsubscript{$\pm$1.12}             & 20.0\textsubscript{$\pm$1.66}             & 30.1\textsubscript{$\pm$1.17}             & 51.4\textsubscript{$\pm$1.36}             & 69.7\textsubscript{$\pm$0.66}             & \underline{-2.36}\textsubscript{$\pm$0.05} \\
    \textbf{SCURank} & \textbf{44.8}\textsubscript{$\pm$0.28}    & \textbf{20.5}\textsubscript{$\pm$0.81}    & \textbf{30.6}\textsubscript{$\pm$0.86}    & 51.7\textsubscript{$\pm$0.78}             & \textbf{70.0}\textsubscript{$\pm$0.27}    & \textbf{-2.34}\textsubscript{$\pm$0.06}    \\
    \midrule\midrule
    \multicolumn{7}{c}{Dataset: BASE}                                                                                                                                                                                                                                                         \\
    \midrule
    UnRank           & 42.3\textsubscript{$\pm$0.11}             & 19.4\textsubscript{$\pm$0.09}             & 29.6\textsubscript{$\pm$0.15}             & 49.8\textsubscript{$\pm$0.36}             & 68.6\textsubscript{$\pm$0.08}             & -2.73\textsubscript{$\pm$0.85}             \\
    \midrule
    ROUGE            & 43.3\textsubscript{$\pm$0.56}             & 20.1\textsubscript{$\pm$0.42}             & 30.3\textsubscript{$\pm$0.42}             & 50.9\textsubscript{$\pm$0.37}             & 69.0\textsubscript{$\pm$0.30}             & -2.62\textsubscript{$\pm$0.02}             \\
    BERTScore        & 43.4\textsubscript{$\pm$0.55}             & 20.3\textsubscript{$\pm$0.48}             & 30.1\textsubscript{$\pm$0.76}             & \underline{51.4}\textsubscript{$\pm$0.30} & 69.0\textsubscript{$\pm$0.37}             & -2.59\textsubscript{$\pm$0.02}             \\
    BLANC            & \underline{43.9}\textsubscript{$\pm$0.68} & \underline{20.6}\textsubscript{$\pm$0.67} & \underline{30.6}\textsubscript{$\pm$0.62} & \underline{51.4}\textsubscript{$\pm$0.93} & \underline{69.5}\textsubscript{$\pm$0.29} & \textbf{-2.45}\textsubscript{$\pm$0.09}    \\
    GPTRank          & 43.0\textsubscript{$\pm$0.35}             & 19.8\textsubscript{$\pm$0.75}             & 29.6\textsubscript{$\pm$0.62}             & 51.2\textsubscript{$\pm$0.81}             & 69.1\textsubscript{$\pm$0.38}             & \underline{-2.52}\textsubscript{$\pm$0.08} \\
    \textbf{SCURank} & \textbf{44.3}\textsubscript{$\pm$0.87}    & \textbf{20.8}\textsubscript{$\pm$0.96}    & \textbf{30.9}\textsubscript{$\pm$1.02}    & \textbf{51.7}\textsubscript{$\pm$0.67}    & \textbf{69.9}\textsubscript{$\pm$0.50}    & \textbf{-2.45}\textsubscript{$\pm$0.10}    \\
    \bottomrule
  \end{tabular}
  }
  \caption{
    Distilled model performance on CNN/DailyMail using the LLMs-9 and BASE \cite{liu-etal-2024-learning} datasets.
    ``LLMs Average'' denotes the mean of nine top LLMs.
    ``UnRank'' is the MLE-only baseline.
    Bold/underline indicate best/second-best results (mean ± std over 5 runs).
  }
  \label{tab:distilled-performance}
\end{table*}

\subsection{Evaluation}
\subsubsection{Test Set}
The reference summaries in the original CNN/DailyMail \citep{nallapati2016abstractive} and XSum \citep{narayan-etal-2018-dont} datasets are known to have quality issues \citep{maynez-etal-2020-faithfulness, kang-hashimoto-2020-improved}, which was also confirmed by \citet{zhang-etal-2024-benchmarking}.
Therefore, we used the human-written references from \citet{zhang-etal-2024-benchmarking} to evaluate the distilled models.
This dataset consists of 54 articles from the CNN/DailyMail dataset and 53 articles from the XSum dataset, each of which contains one to three human-written references.
If a summary has multiple references, the highest score is selected.

\subsubsection{Metrics}
We use ROUGE-1, ROUGE-2, and ROUGE-L \citep{lin-2004-rouge} as our primary evaluation metrics.
To complement these lexical metrics, we additionally include three model-based metrics: BERTScore \citep{zhang2019bertscore}, BLEURT \citep{sellam-etal-2020-bleurt}, and BARTScore \citep{yuan2021bartscore}, which capture semantic similarity using pre-trained language models.

\section{Results}

\begin{table*}[t]
  \centering
  \resizebox{0.92\textwidth}{!}{
  \begin{tabular}{l|cccccc}
    \toprule
                     & ROUGE-1                                   & ROUGE-2                                   & ROUGE-L                                   & BLEURT                                    & BERTScore                                 & BARTScore                                  \\
    \midrule\midrule
    \multicolumn{7}{c}{LLMs Results}                                                                                                                                                                                                                                                          \\
    \midrule
    GPT-4o           & 43.2                                      & 16.1                                      & 29.3                                      & 55.1                                      & 70.3                                      & -2.85                                      \\
    Gemini-1.5-pro   & 43.9                                      & 16.1                                      & 30.5                                      & 55.6                                      & 71.3                                      & -2.65                                      \\
    Mistral-Large    & 42.8                                      & 15.8                                      & 29.1                                      & 54.1                                      & 69.7                                      & -2.82                                      \\
    LLMs Average     & 42.5                                      & 15.9                                      & 29.4                                      & 52.9                                      & 69.6                                      & -2.91                                      \\
    \midrule\midrule
    \multicolumn{7}{c}{Dataset: LLMs-9}                                                                                                                                                                                                                                                       \\
    \midrule
    UnRank           & \underline{45.3}\textsubscript{$\pm$0.11} & \underline{19.2}\textsubscript{$\pm$0.23} & 31.5\textsubscript{$\pm$0.24}             & 54.2\textsubscript{$\pm$0.14}             & 70.6\textsubscript{$\pm$0.09}             & -2.71\textsubscript{$\pm$0.01}             \\
    \midrule
    ROUGE            & 43.9\textsubscript{$\pm$1.05}             & 18.9\textsubscript{$\pm$0.72}             & \textbf{32.5}\textsubscript{$\pm$0.49}    & \underline{54.4}\textsubscript{$\pm$0.55} & \textbf{71.1}\textsubscript{$\pm$0.14}    & \textbf{-2.47}\textsubscript{$\pm$0.04}    \\
    BERTScore        & 44.0\textsubscript{$\pm$1.04}             & 18.5\textsubscript{$\pm$0.46}             & 31.8\textsubscript{$\pm$0.31}             & 53.6\textsubscript{$\pm$0.66}             & 70.5\textsubscript{$\pm$0.17}             & \underline{-2.52}\textsubscript{$\pm$0.06} \\
    BLANC            & 44.3\textsubscript{$\pm$0.52}             & 18.8\textsubscript{$\pm$0.39}             & 31.7\textsubscript{$\pm$0.32}             & 54.1\textsubscript{$\pm$0.92}             & 70.3\textsubscript{$\pm$0.22}             & -2.55\textsubscript{$\pm$0.02}             \\
    GPTRank          & 44.2\textsubscript{$\pm$1.67}             & 18.4\textsubscript{$\pm$1.17}             & 31.1\textsubscript{$\pm$1.16}             & 54.2\textsubscript{$\pm$1.52}             & 70.3\textsubscript{$\pm$0.58}             & -2.57\textsubscript{$\pm$0.04}             \\
    \textbf{SCURank} & \textbf{45.4}\textsubscript{$\pm$0.26}    & \textbf{19.3}\textsubscript{$\pm$0.23}    & \underline{32.0}\textsubscript{$\pm$0.26} & \textbf{55.4\textsubscript{$\pm$0.44}}    & \underline{70.8}\textsubscript{$\pm$0.25} & -2.54\textsubscript{$\pm$0.02}             \\
    \bottomrule
  \end{tabular}
  }
  \caption{
    Distilled model performance on XSum using the LLMs-9 datasets.
    ``LLMs Average'' denotes the mean of nine top LLMs.
    ``UnRank'' is the MLE-only baseline.
    Bold/underline indicate best/second-best results (mean $\pm$ std over 10 runs).
  }
  \label{tab:xsum-d-performance}
\end{table*}

\subsection{Main Results}
\label{sec:distilled-performance}
\subsubsection{SCURank vs. Baselines}
Table~\ref{tab:distilled-performance} shows the performance of the distilled models trained using different ranking methods on CNN/DailyMail.
The distilled model trained with SCURank outperforms those trained with the other ranking methods, achieving the highest scores in all metrics, except BLEURT in the LLMs-9 dataset.
To further evaluate the effectiveness of SCURank, we test the distilled model on the XSum version of LLMs-9.
Different from CNN/DailyMail, we conduct 10 runs on the XSum dataset (instead of 5) to ensure more robust evaluation.
In Table~\ref{tab:xsum-d-performance}, the model trained with SCURank  achieves the best scores in ROUGE-1, ROUGE-2, and BLEURT, and the second-highest scores in ROUGE-L and BERTScore.
Interestingly, the model trained with the ROUGE-based ranking method also performs well on XSum, achieving the highest scores in ROUGE-L, BERTScore, and BARTScore; however, its scores on other metrics are notably lower than those of the SCURank model.
Overall, the results on both datasets suggest that SCURank is a more robust and effective ranking method than GPTRank or traditional automatic metrics.
The significance test results between SCURank and GPTRank are provided in Appendix~\ref{sec:statistical_test}. % Appendix 

\subsubsection{MLE vs. Contrastive Learning}
We also evaluate the UnRank model, which is trained using maximum likelihood estimation (MLE) on summaries generated by LLMs.
In Table~\ref{tab:distilled-performance}, the UnRank model trained on the LLMs-9 dataset achieves the highest score in ROUGE-2 and BLEURT, and the second-highest in BERTScore.
However, the UnRank model's performance is comparatively lower when trained on the BASE dataset, suggesting that the outputs  of multiple LLMs can have a positive effect through MLE training.
Additionally, the UnRank model trained on the XSum dataset performs worse across all metrics compared to the distilled model trained with contrastive learning (SCURank), as shown in Table~\ref{tab:xsum-d-performance}.
These findings indicate that contrastive learning, especially with SCURank, offers more effective supervision than MLE alone.
The case studies in Appendix~\ref{sec:scurank-vs-mle} further illustrate the advantage of SCURank over MLE-only training.

\subsubsection{Comparison with LLM Outputs}
Both Table~\ref{tab:distilled-performance} and Table~\ref{tab:xsum-d-performance} show that
the distilled models consistently outperform the LLMs average in all metrics except BLEURT.
However, certain individual LLMs, such as Mistral-Large, still achieve higher scores than the distilled model in ROUGE-1, BLEURT, and BERTScore.
These results suggest that the performance of the distilled models have reached a performance level comparable to top-performing LLMs, indicating the effectiveness of the distillation process with ranking-based contrastive learning.
More scores of individual LLMs are provided in the Appendix~\ref{sec:llms-result}. % Appendix

\subsection{Stability of Ranking}
To assess the consistency of ranking approaches, we evaluate SCURank and GPTRank on the LLMs-9 dataset (CNN/DailyMail).
We perform five independent ranking runs on 1,000 samples. % and report the stability of the resulting rankings.

\begin{table}[!htb]
  \centering
  \resizebox{0.8\columnwidth}{!}{
  \begin{tabular}{l|ccc|c}
    \toprule
             & $\tau$        & $\rho$        & $r$           & $\alpha$      \\
    \midrule\midrule
    GPTRank  & 76.8          & 78.4          & 78.4          & 96.4          \\
    GPTRank* & 16.7          & 22.4          & 22.4          & 3.0           \\
    SCURank  & \textbf{66.1} & \textbf{72.0} & \textbf{72.0} & \textbf{84.6} \\
    % SCURank-umap & 50.9	& 60.2 & 60.2 &	64.1 \\ 
    \bottomrule
  \end{tabular}
  }
  \caption{
    Stability evaluation of ranking methods.
    * indicates that summaries were shuffled before ranking (SCURank is order-invariant).
    Metrics: Kendall's $\tau$, Spearman's $\rho$, Pearson's $r$, and Krippendorff's $\alpha$.
  }
  \label{tab:stability}
\end{table}

For each method, we adopt a representative-ranking strategy:
for each sample, we select the run with the highest average correlation with the other runs as the representative, reporting its mean correlation.
Rankings are evaluated using Kendall's $\tau$, Spearman's $\rho$, and Pearson's $r$, along with Krippendorff's $\alpha$ to measure inter-run agreement, where values above 0.8 are generally considered to indicate reliable agreement.

As shown in Table~\ref{tab:stability}, GPTRank exhibits high correlation when the input summary order is fixed, but its performance degrades substantially under randomized ordering.
This behavior is consistent with prior findings that LLM-based ranking methods are sensitive to input order.

In contrast, SCURank achieves consistently high reliability across runs, with a Krippendorff's $\alpha$ of 0.846, exceeding the commonly accepted threshold of 0.8.
This robustness stems from using LLMs solely for SCU extraction rather than direct ranking, thereby avoiding order sensitivity.
Overall, these results demonstrate that SCURank is more stable and robust than GPTRank.

\begin{figure}[htb]
  \centering
  \includegraphics[width=\linewidth]{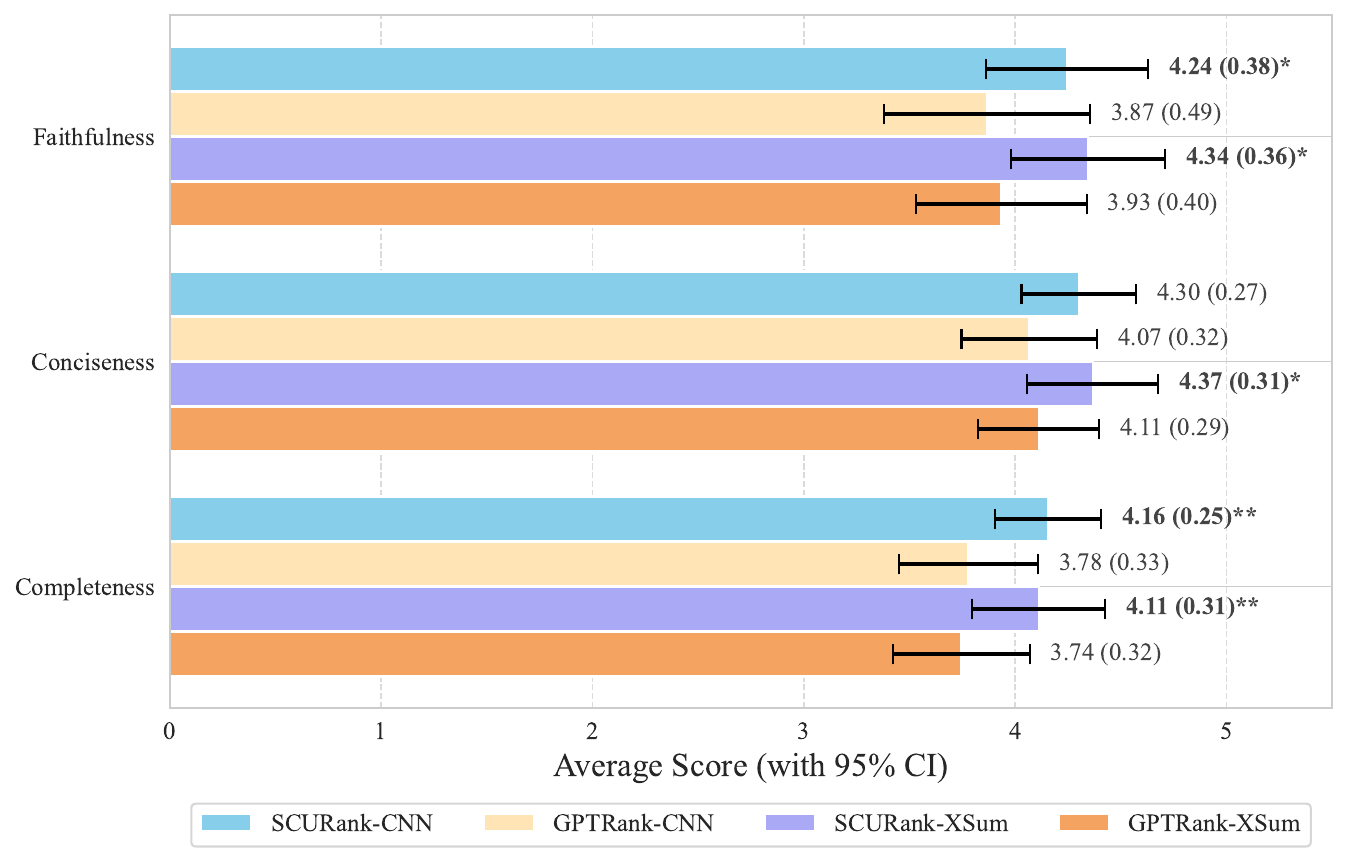}
  \caption{
    Human evaluation results comparing distilled models trained with SCURank and GPTRank.
    Each summary was scored by 3 annotators.
    Significance (paired t-test): * $p < 0.05$, ** $p < 0.01$.
  }
  \label{fig:human-comparison}
\end{figure}

\begin{figure}[ht]
  \centering
  \includegraphics[width=\linewidth]{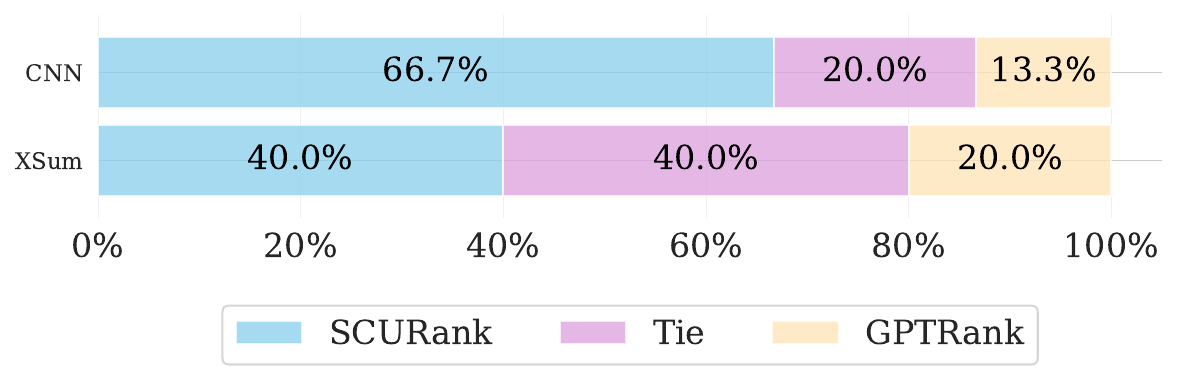}
  \caption{
    Human preference evaluation results between distilled models trained with SCURank and GPTRank.
  }
  \label{fig:human-preference}
\end{figure}

\subsection{Human Evaluation}
\label{sec:human-eval}
To ensure our results align with human perception, we conducted a human evaluation to compare the summaries generated by distilled models trained with SCURank and GPTRank.
We randomly sampled 30 articles from the test sets of CNN/DailyMail and XSum, respectively.
For each article, we presented the source text and pairs of summaries generated by the two models to three independent annotators recruited via Amazon Mechanical Turk (MTurk).
Annotators were asked to assess the summaries in two ways: (1) scoring them on a Likert scale (1-5) across three dimensions: \textit{Faithfulness}, \textit{Conciseness}, and \textit{Completeness}; and (2) providing an overall preference (Win/Tie/Loss).
The details of the quality of annotators and evaluation criteria are provided in the Appendix~\ref{sec:human-eval-impl}.

Figure~\ref{fig:human-comparison} presents the average scores for each metric.
The distilled model trained with SCURank consistently outperforms the GPTRank baseline across all dimensions on both datasets.
Notably, SCURank demonstrates highly significant improvements ($p < 0.01$) in \textit{Completeness} for both CNN/DailyMail and XSum, suggesting that our SCU-based approach effectively encourages the model to retain more key information.

Figure~\ref{fig:human-preference} illustrates the pairwise preference results.
On both datasets, SCURank exhibits a dominant advantage over GPTRank.
On the XSum dataset, while the percentage of ties increases to 40.0\% (likely due to the shorter length of XSum summaries making distinct differentiation harder), SCURank still achieves a 40.0\% win rate, double that of GPTRank (20.0\%).
These results indicate that SCURank produces more faithful, concise, complete, and overall better summaries under the human evaluations.
To further illustrate the qualitative differences, we provide the sample summaries in Appendix \ref{sec:scurank-vs-gptrank}.

\begin{figure*}[ht]
  \centering
  \includegraphics[width=\linewidth]{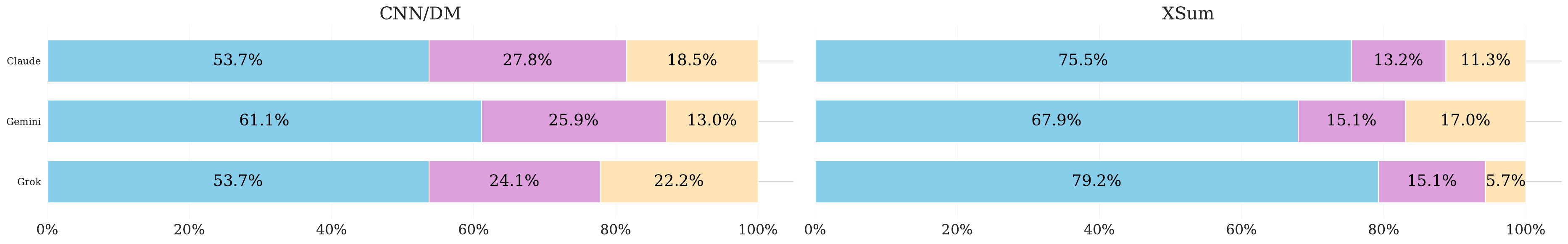}
  \caption{
    Win-tie-loss rates between distilled models trained with SCURank and GPTRank, evaluated by three LLM judges: Claude-4-sonnet, Gemini-2.5-pro, and Grok 4.
    % The dataset is identical to the one cited in the Test Set, comprising 54 articles from CNN/DailyMail and 53 articles from XSum.
  }
  \label{fig:distilled-model-comparison-llm}
\end{figure*}

\subsection{LLM-based Evaluation}
\label{sec:llm-based-eval}
In order to further validate the effectiveness of SCURank, we evaluate the distilled models trained with candidate summaries ranked by SCURank and GPTRank, using three of the latest LLMs: Grok-4 \citep{xai2025}, Claude-4-sonnet \citep{claude2025}, and Gemini-2.5-pro \citep{gemini_models}.
For each test sample, we present the summaries generated by the two distilled models (trained using SCURank and GPTRank rankings, respectively) in the prompt of an LLM, asking it to decide which summary is better.
To reduce the positional bias, each pairwise comparison is conducted twice, with the order of the summaries reversed in the second round.
A model receives a point if it wins two rounds, or if it wins one round and ties in the other.

The results are shown in Figure~\ref{fig:distilled-model-comparison-llm}.
The distilled model trained with SCURank consistently outperforms the one trained with GPTRank across both datasets.
The results further confirm that the ranking superiority of SCURank as a ranking method for training summarization models.

\subsection{Writing Style}
To investigate the impact of distillation from multiple LLMs, we analyze the abstractiveness of summaries generated by distilled models trained on BASE and LLMs-9.

Following \citet{grusky-etal-2018-newsroom}, we measure abstractiveness using coverage and density, where lower values indicate more abstractive summaries.
Coverage measures the proportion of words in the summary that appear in the source document, while density measures the average length of contiguous spans shared between them.

Table~\ref{tab:writing-style} compares distilled models with human-written summaries \citep{zhang-etal-2024-benchmarking} and LLM-generated summaries.
The distilled model trained on LLMs-9 consistently achieves lower coverage and density than the one trained on BASE, indicating improved abstractiveness.

Although distilled models still exhibit higher coverage and density than both LLMs and human-written summaries, the gap is substantially reduced.
These results suggest that distillation from diverse LLM-generated summaries enhances abstractiveness and moves the model closer to human writing style.
This improvement likely stems from the increased diversity of training summaries, since different LLMs often express similar content using different phrasings.
Learning from multiple LLMs therefore encourages the model to capture semantic content rather than copying spans from the source text.

\begin{table}[htb]
  \centering
  \resizebox{0.9\columnwidth}{!}{
  \begin{tabular}{l|cc|cc}
    \toprule
              & \multicolumn{2}{c|}{Coverage} & \multicolumn{2}{c}{Density}                 \\
    \midrule
              & BASE                          & LLMs-9                      & BASE  & LLMs-9   \\
    \midrule\midrule
    Human     & --                            & 0.81                        & --    & 2.07  \\
    LLMs      & --                            & 0.83                        & --    & 2.70  \\
    \midrule
    SCURank   & 0.95                          & 0.94                        & 10.34 & 6.73  \\
    ROUGE     & 0.98                          & 0.97                        & 18.46 & 16.34 \\
    BERTScore & 0.97                          & 0.95                        & 17.31 & 9.99  \\
    BLANC     & 0.96                          & 0.94                        & 11.82 & 7.56  \\
    GPTRank   & 0.96                          & 0.93                        & 14.80 & 5.97  \\
    Average   & 0.96                          & 0.95                        & 14.54 & 9.32  \\
    \bottomrule
  \end{tabular}
  }
  \caption{
    Coverage and density of summaries generated by models trained on the BASE and LLMs-9 datasets on CNN/DailyMail.
    For reference, we also report scores for human-written summaries \cite{zhang-etal-2024-benchmarking} and LLM-generated candidates (LLMs).
    \textbf{Lower} scores indicate higher abstractiveness.
  }
  \label{tab:writing-style}
\end{table}

\section{Conclusions}
We propose SCURank, an information-based ranking method that leverages Summary Content Units (SCUs) to provide more stable and semantically meaningful ranks than existing approaches.
Experimental results demonstrate that SCURank improves contrastive learning performance, outperforming both GPTRank and traditional automatic metrics.
Moreover, training with diverse summaries generated by multiple LLMs has been shown to enhance abstractiveness while maintaining overall performance.
These findings highlight SCURank as an effective solution to rank high-quality summaries and demonstrate the potential of multi-LLM distillation for enhanced abstractive summarization.

\section*{Limitations}
In SCUs extraction, we used the gpt-4o-mini with 1-shot examples from REALSumm due to budget constraints.
The results of the intrinsic evaluation showed that the SCUs extracted by gpt-4o with 3-shot examples from REALSumm achieved the highest score.
Moreover, if the total number of SCUs is too small, the clustering may change the parameters of the HDBSCAN.

The SCURank should be more competitive with more candidate summaries, but this study targets the comparison with GPTRank \citep{liu-etal-2024-learning}, which uses one reference and eight candidate summaries for training with contrastive learning.

\section*{Acknowledgements}
This work was supported by the National Science and Technology Council, Taiwan, under Grants NSTC 114-2223-E-007-011 and NSTC 114-2222-E-182-001-MY2.
We would like to express our sincere gratitude to the reviewers for their thoughtful comments and valuable feedback.

% Bibliography entries for the entire Anthology, followed by custom entries
%\bibliography{custom,anthology-overleaf-1,anthology-overleaf-2}

% Custom bibliography entries only
\bibliography{customv2}

\appendix

% \section{Example Appendix}
% \label{sec:appendix}

% This is an appendix.
\section{Details of LLM-generated Summaries}
\label{sec:llm-gen-details}
We adopt the following nine LLMs to generate summaries for both CNN/DailyMail and XSum: 
GPT-4 \citep{openai2024gpt4technicalreport},
GPT-4o \citep{openai2024gpt4ocard},
GPT-4o-mini \citep{gpt4omini},
Gemini-1.5-flash,
Gemini-1.5-pro \citep{gemini_models},
Llama-3.1-instruct-70b,
Llama-3.1-instruct-405b \citep{llama3modelcard},
Claude-3.5-sonnet \citep{claude2024},
and Mistral-Large-2407 \citep{mistral2024}.

The prompt is designed to generate brief summaries for the CNN/DailyMail-type datasets (BASE and LLMs-9, as detailed in the primary paper).
The feature of the CNN/DailyMail dataset, in which the summary is constructed in three sentences, is maintained.
\begin{tcolorbox}[colback=gray!10, colframe=gray!50, boxrule=0.8pt, width=\columnwidth]
\small
  Summarize the main content of the following news article in three sentences.
\end{tcolorbox}

\paragraph{XSum Summary Generation}
The following prompt is used for the summary generation task for XSum-type datasets.
Although XSum summaries are typically one sentence long, we modify the prompt to request a three-sentence summary.
This design choice ensures that the distillation process yields a model capable of generating summaries comparable in length to those in the test set.
\begin{tcolorbox}[colback=gray!10, colframe=gray!50, boxrule=0.8pt, width=\columnwidth]
\small
    Summarize the following article in three sentences. Ensure the summary is concise, with a total word count between 40 and 50 words.
\end{tcolorbox}

\section{Details of Summary Content Units}\label{sec:scu-details}
\subsection{Summary Content Units Extraction}
\label{sec:scu-prompts}
Below is the prompt used for the SCU extraction task.
\begin{tcolorbox}[colback=gray!10, colframe=gray!50, boxrule=0.8pt, width=\columnwidth]
\small
  You split the provided input in small sentences separated by an \#. The split sentences represent subsentences of the original sentences. \\
Example inputs:\\
Anuradha Koirala and 425 young women and girls have been sleeping outdoors because of aftershocks. Pushpa Basnet and 45 children she cares for were forced to evacuate their residence. Seven other CNN Heroes and their organizations now assisting in relief efforts. \\
Example outputs: \\
Anuradha Koirala has been sleeping outdoors. \# 425/many young women and girls have been sleeping outdoors. \# Many people have been sleeping outdoors because of aftershocks. \# Pushpa Basnet was forced to evacuate her residence. \# Pushpa Basnet cares for 45 children. \# The children were forced to evacuate their residence. \# Anuradha Koirala was a CNN Hero. \# Pushpa Basnet was a CNN Hero. \# Seven other CNN Heros were now assisting relief efforts. \# The organizations of CNN Heros were now assisting relief efforts.
\end{tcolorbox}

\subsection{SCUs Intrinsic Evaluation}
\label{sec:scu-intrinsic-evaluation}
Since we changed both the LLM and the instruction for SCU extraction, we conducted an intrinsic evaluation to assess the quality of the extracted SCUs.
This intrinsic evaluation follows the methodology of \citet{nawrath-etal-2024-role}, and is performed on the REALSumm dataset \citep{bhandari-etal-2020-evaluating} and the PyrXSum dataset \citep{zhang-bansal-2021-finding}.
The evaluation score is built by iterating over each pair of human-annotated SCUs and the llm-extracted SCUs, averaging the maximum ROUGE-1-F1 score obtained for each human-annotated SCU.
As this metric is recall-biased, we additionally compute the score in the reverse direction, following \citet{nawrath-etal-2024-role}.

Table~\ref{tab:intrinsic-results} presents SCUs extraction quality, with scores generally increasing as the number of examples increases.
Across both REALSumm and PyrXSum, GPT-4o and GPT-4o-mini outperform GPT-3.5-turbo.
On REALSumm, GPT-4o achieves the highest score (0.79), slightly higher than GPT-4o-mini (0.76).
On PyrXSum, both GPT-4o and GPT-4o-mini reach similar, high quality (0.75).
All models exhibit substantial improvement when using a 1-shot example compared to zero-shot setting.

\textbf{R} represent the maximum ROUGE-1-F1 score found for each human-annotated SCU, while \textbf{P} represent the maximum ROUGE-1-F1 score found for each llm-extracted SCU.

\begin{table}[ht]
  \centering
  \begin{tabular}{l|cc|cc}
    \toprule
    \multirow{2}{*}{} & \multicolumn{2}{c}{RealSumm} & \multicolumn{2}{c}{PyrXSum} \\ 
    \cmidrule(lr){2-3} \cmidrule(lr){4-5}
    & R & P & R & P \\
    \midrule
    \midrule
    SGUs\_3.5 & .58 & .67 & .58 & .63 \\
    SGUs\_4 & .61 & .69 & .61 & .66 \\
    \midrule
    \multicolumn{5}{l}{gpt-3.5-turbo} \\
    0-shot & .54 & .66 & .55 & .62 \\
    1-shot & .62 & .74 & .65 & .73 \\
    3-shot & .68 & .78 & .70 & .74 \\
    \midrule
    \multicolumn{5}{l}{gpt-4o-mini} \\
    0-shot & .55 & .67 & .58 & .67 \\
    1-shot & .70 & .76 & .72 & .72 \\
    3-shot & .76 & .79 & .75 & .77 \\
    \midrule
    \multicolumn{5}{l}{gpt-4o} \\
    0-shot & .51 & .59 & .60 & .67 \\
    1-shot & .73 & .78 & .71 & .76 \\
    3-shot & .77 & .79 & .75 & .77 \\
    \bottomrule
  \end{tabular}
  \caption{
    Intrinsic evaluation results on REALSumm and PyrXSum.
    Results for SGUs\_3.5 and SGUs\_4 are taken from the original paper.
    Each block reports results for a specific LLM and number of shots.
    \textbf{R} and \textbf{P} denote the maximum ROUGE-1 F1 scores in the two evaluation directions.
  }
  \label{tab:intrinsic-results}
\end{table}

\section{Analysis of Scoring and Length Penalty}
\label{sec:length_penalty_analysis}

To mitigate potential length bias in SCURank, we evaluate the relationship between summary length ($||s_i||$) and various scoring configurations.
Our goal is to minimize the correlation between the final rank and word count to ensure the model prioritizes information richness over verbosity.

\subsection{Evaluation of Length Penalty}
We use Kendall's $\tau$ to measure the correlation between summary scores and lengths.
As shown in Table~\ref{tab:penalty_test}, raw SCU counts ($|c|$) without normalization (N/A) exhibit a strong positive correlation with length across both CNN/DailyMail and XSum datasets.
While a linear penalty ($\text{len}(s)$) leads to over-penalization (negative correlation), the \textbf{square root penalty ($\sqrt{\text{len}(s)}$)} consistently achieves a correlation closest to zero, effectively decorrelating the score from summary length.

\subsection{Justification of Linear Sum}
We further compared the linear sum against square root and logarithmic transformations of the raw SCU scores.
The results indicate that more complex scoring functions do not significantly improve length decorrelation.
In the CNN/DM dataset, the simple linear sum ($|c|$) paired with a square root penalty achieves the lowest correlation (0.1316).
Following the principle of simplicity, we adopt the linear sum as the primary scoring mechanism defined in Eq.~\ref{eq:SCUScore}.

\begin{table}[ht]
\centering
\small
\resizebox{\columnwidth}{!}{
\begin{tabular}{llcccc}
\toprule
\textbf{Dataset} & \textbf{Score Type} & \textbf{N/A} & \textbf{len($s$)} & \textbf{$\sqrt{\text{len}(s)}$} & \textbf{$\log(\text{len}+1)$} \\
\midrule
\multirow{3}{*}{CNN/DM} & Sum ($|c|$)           & 0.3666 & -0.1394 & \textbf{0.1316} & 0.2931 \\
                     & $\sqrt{|c|}$     & 0.4230 & -0.1609 & 0.1412 & 0.3022 \\
                     & $\log(|c|+1)$   & 0.4293 & -0.1595 & 0.1546 & 0.3202 \\
\midrule
\multirow{3}{*}{XSUM} & Sum ($|c|$)           & 0.2091 & -0.2272 & \textbf{-0.0346} & 0.0768 \\
                     & $\sqrt{|c|}$     & 0.2761 & -0.2707 & -0.0135 & 0.1044 \\
                     & $\log(|c|+1)$   & 0.2705 & -0.2752 & -0.0047 & 0.1070 \\
\bottomrule
\end{tabular}
}
\caption{Kendall's $\tau$ correlation between various scoring functions and summary word count. Values closer to zero indicate superior mitigation of length bias. Bold values denote the settings used in SCURank.}
\label{tab:penalty_test}
\end{table}

\section{Implementation Details}
\label{sec:implementation-details}
\subsection{Distillation in CNN/DailyMail}
We employed the BART model \citep{lewis-etal-2020-bart} as the target model, initialized with the checkpoint of facebook/bart-large-cnn\footnote{https://huggingface.co/facebook/bart-large-cnn}.
Prior to the fine-tuning stage described in \citet{liu-etal-2024-learning}, a warm-up training using maximum likelihood estimation (MLE) was conducted with 10,000 GPT-3.5 summaries, provided by \citet{liu-etal-2024-learning}.
Then, the process continued with contrastive learning, as detailed in Section: \textbf{Contrastive Learning} of our main paper.
To ensure reliability, each experiment was repeated five times, and the average performance was reported.

\subsubsection{Explanation to the UnRank Baseline}
To evaluate the effectiveness of contrastive learning, we conducted an additional experiment using MLE training only after the warm-up stage.
In this setting, the distilled model was trained on all pairs of summaries in the dataset, processing nine different summaries per document.
We then compared its performance with that of the model trained with contrastive learning.
In our experiments, we referred to the model trained with only MLE as "UnRank."

\subsection{Distillation in XSum}
The training process for CNN/DailyMail and XSum follows the same procedure.
The only difference is the training size of LLMs-9 for the XSum dataset.
We used 9,000 gpt-4o-mini summaries to fine-tune the BART checkpoint for the original XSum dataset\footnote{https://huggingface.co/facebook/bart-large-xsum}.
The number of summaries is fewer than 10,000 due to the limited size of the original XSum dataset.

\subsection{HDBSCAN Clustering}
For each SCUs clustering, since the number of samples is relatively small, we set the minimum cluster size and the minimum samples size jointly to \textbf{2}, which allows us to capture even small clusters of semantically similar SCUs.
Furthermore, the cluster selection epsilon is set to \textbf{0.15} to stabilize the clustering results, reducing over-fragmentation caused by small local density variations.
To ensure that all SCUs are included in the ranking, we treat `noise' outliers as individual clusters.

% \subsection{Stability of Ranking}
% \label{sec:stability-impl}
% \subsubsection{Repeated Ranking Procedure}
% For each ranking method, we conduct five independent runs on the same set of 1,000 samples.
% Each run produces a complete ranking of the candidate summaries for each sample.

% To summarize ranking consistency, we adopt a representative-ranking strategy.
% For each sample, we iteratively select one run as the reference ranking and compute its correlations with the remaining runs.
% The ranking with the highest average correlation is selected as the representative, and the corresponding mean correlation score is reported.

% \subsubsection{Correlation and Reliability Metrics}
% We measure ranking correlations using Kendall's $\tau$, Spearman's $\rho$, and Pearson's $r$.
% In addition, we compute Krippendorff's $\alpha$ to quantify inter-run agreement, where values above 0.8 are generally considered to indicate reliable agreement.

\section{Dataset Statistics}
\label{sec:data-stat}
Two datasets, BASE \cite{liu-etal-2024-learning}, LLMs-9, were employed to train the distilled model.
Table~\ref{tab:data-statistic} shows the statistics of these datasets.
\begin{table}[ht]
  \centering
  \resizebox{0.9\columnwidth}{!}{
  \begin{tabular}{l|cc|cc}
    \toprule
    \multirow{2}{*}{Dataset}
    & \multicolumn{2}{c}{\#Examples} & \multicolumn{2}{c}{Avg. Words} \\
    & Train & Valid & Doc. & Sum. \\
    \midrule
    BASE (CNN/DM) & 1k & 100 & 601.7 & 75.7\\
    LLMs-9 (CNN/DM) & 1k & 100 & 601.7 & 85.7\\
    LLMs-9 (XSum) & 1k & 100 & 423.5 & 52.4\\
    \bottomrule
  \end{tabular}
  }
  \caption{
    Dataset statistics used in this study.
    The \#Example and Avg. Words columns report the number of examples and average number of words in documents and summaries, respectively.
    The BASE dataset is provided by \citet{liu-etal-2024-learning}, where the articles drawn from CNN/DailyMail (CNN/DM).
  }
  \label{tab:data-statistic}
\end{table}

\section{Significant Test Evaluation}
\label{sec:statistical_test}
\subsection{Paired Bootstrap Test}
To evaluate the statistical significance of the performance difference between SCURank and GPTRank, we conducted a paired bootstrap test on the evaluation metrics reported in Table~\ref{tab:distilled-performance} and Table~\ref{tab:xsum-d-performance}.
For each metric, we collected all scores for the samples in the evaluation set in each run, resulting in two sets of scores for SCURank and GPTRank.
We then performed 10,000 bootstrap resamples of these score pairs, calculating the mean difference between SCURank and GPTRank for each resample.
The p-value was computed as twice the proportion of bootstrap samples on the minority side of zero (two-tailed test).
A p-value less than 0.05 was considered statistically significant, and is denoted with an asterisk (*) in the results. 

\begin{table}[ht]
  \centering
  \resizebox{\columnwidth}{!}{
    \begin{tabular}{lcccccc}
      \toprule
      \textbf{Data Type} & \textbf{R\_1} & \textbf{R\_2} & \textbf{R\_L} & \textbf{BLEU} & \textbf{BScore} & \textbf{BaScore} \\
      \midrule
      BASE          & 0.012* & 0.035* & 0.013* & 0.186  & 0.017* & 0.195 \\
      LLMs-9        & 0.043* & 0.317  & 0.311  & 0.448  & 0.255  & 0.699 \\
      LLMs-9 (XSum) & 0.000* & 0.002* & 0.000* & 0.000* & 0.001* & 0.000* \\
      \bottomrule
    \end{tabular}
    }
  \caption{
    Paired bootstrap test $p$-values.
    * denotes SCURank significantly outperforms GPTRank ($p < 0.05$).
    Metrics are abbreviated (R: ROUGE, BScore: BERTScore, BaScore: BARTScore).
  }
  \label{tab:evaluation_metrics}
\end{table}

\subsection{Analysis of Significance Results}
Table~\ref{tab:evaluation_metrics} reveals distinct patterns across datasets.
On XSum (LLMs-9), SCURank achieves highly significant improvements ($p < 0.01$) across all automatic metrics, demonstrating its effectiveness in highly abstractive summarization scenarios.
On CNN/DM (BASE), significant improvements ($p < 0.05$) are observed in ROUGE-1, ROUGE-2, ROUGE-L, and BERTScore, further confirming the superiority of our method.
For CNN/DM (LLMs-9), although SCURank consistently achieves higher mean scores than the baseline across several metrics, the differences do not reach statistical significance.
This does not imply that SCURank is less effective in this setting; when model outputs are already of high quality, automatic metrics are known to have limited sensitivity in detecting subtle but meaningful improvements.

This is further supported by human evaluation (Section~\ref{sec:human-eval}), where SCURank shows significant gains in Faithfulness ($p < 0.05$) and Completeness ($p < 0.01$), consistent with our LLM-based evaluation (Section~\ref{sec:llm-based-eval}).

\section{Human Evaluation}
\label{sec:human-eval-impl}
\subsection{Guidelines}
The task was hosted on Amazon Mechanical Turk (MTurk).
Each task included an article and two summaries generated by the distilled models trained with SCURank and GPTRank.
To mitigate positional bias, the summaries were randomly ordered.
Three independent annotators were recruited for each task to ensure reliable evaluation.

\begin{figure*}[t]
  \centering
  \includegraphics[width=\linewidth]{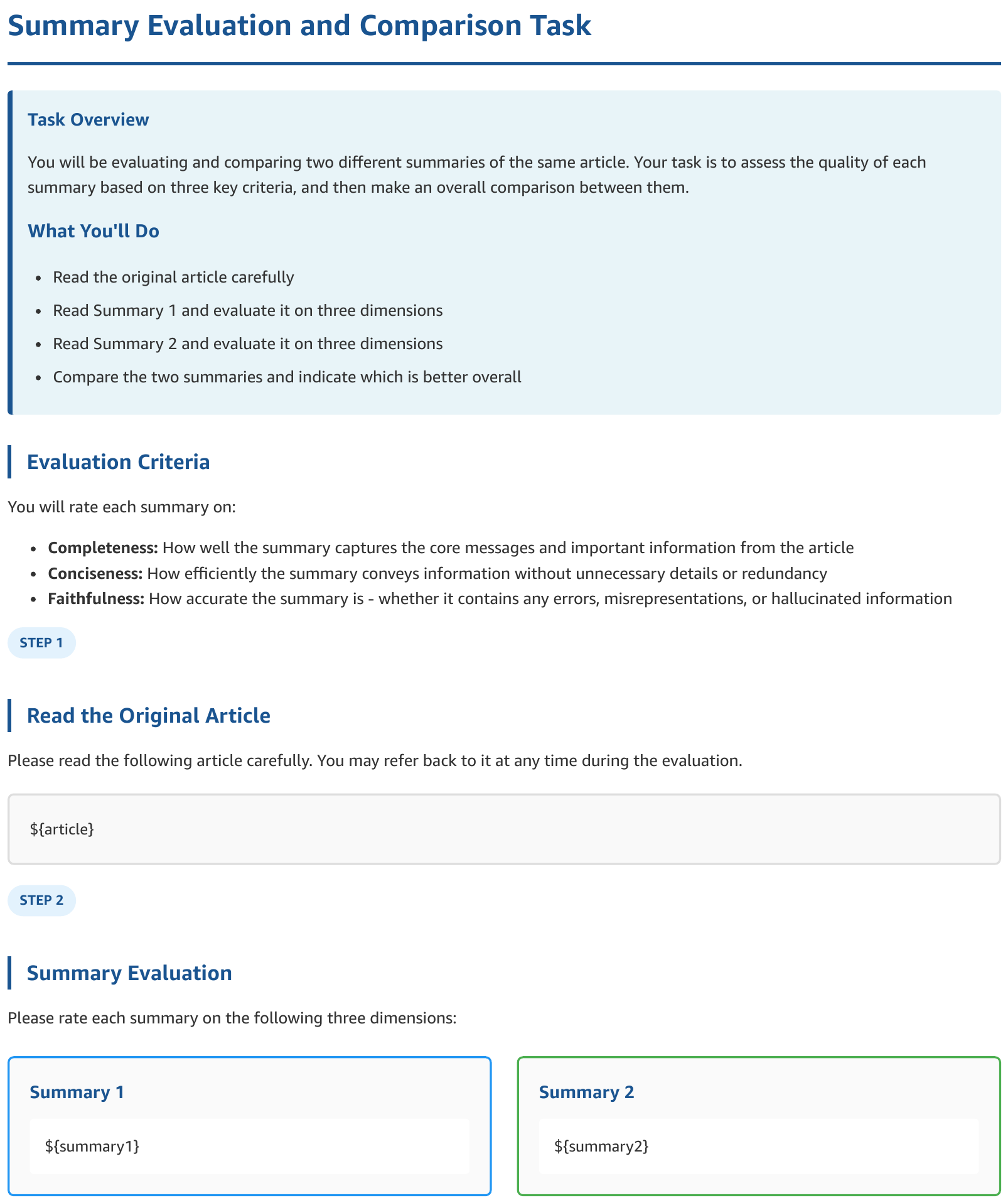}
  \caption{Annotation guidelines provided to annotators for human evaluation on MTurk.}
  \label{fig:human-eval-guidelines}
\end{figure*}

Under the previous qualification criteria, annotators were required to complete a minimum of 500 HITs, maintain an approval rate of at least 90\%, and use English as their primary language.
Each annotator was provided with detailed guidelines explaining the evaluation criteria and scoring system.
The guidelines are defined in Figure~\ref{fig:human-eval-guidelines}.

\subsection{Evaluation Criteria}
Each summary is evaluated based on three dimensions: Faithfulness, Conciseness, and Completeness.
Then, annotators rate each criterion on a five-point Likert scale, where one indicates poor quality and five indicates excellent quality.
Finally, annotators are asked to compare the two summaries and select which one they prefer overall.
A summary needs to receive at least two votes to be considered the preferred summary.
If both summaries receive one vote each and one tie, it is considered a tie.

\section{Sample Summaries}
\subsection{SCURank vs. MLE}
\label{sec:scurank-vs-mle}
To evaluate the performance of the SCURank distilled model and the MLE distilled model, we provide sample summaries generated by both models. 
Table~\ref{tab:sample-summaries} presents these summaries alongside the human-written summaries.
The SCURank distilled model explicitly mentions ``amid population growth and environmental pressures'' as the primary cause of the Hainan gibbons' decline.
Similarly, human-written summary 1 conveys the same idea but with different wording, ``due to deforestation and human population growth.''
This differentiate in phrasing results in a lower ROUGE-1 score for the SCURank distilled model in this case.
In contrast, the MLE distilled model focuses primarily on the role of historical Chinese documents, which is similar to the human-written summary 3.
This similarity leads to a higher ROUGE-1 score for the MLE distilled model.
This result indicates that, while the SCURank distilled model provides more comprehensive information, its summary may sometimes yield a lower ROUGE-1 score due to differences in content focus.
This suggests that the ROUGE-n scores reflect only content similarity rather than overall summary quality.

\label{sec:sample-summaries}
\subsection{SCURank vs. GPTRank}
\label{sec:scurank-vs-gptrank}
To further demonstrate the capability of SCURank in capturing diverse information, we present a qualitative comparison between SCURank and GPTRank in Table \ref{tab:case_study_scurank_vs_gptrank}.
In this example, the source article describes a controversy involving the partner of a political leader.

As shown in the table, the \textbf{GPTRank} model tends to focus heavily on sensational and high-frequency keywords found in the lead paragraph (e.g. specific quotes about ``rape fantasies'' and ``complex human beings'').
While these details are salient, the summary misses the narrative context of their relationship.

In contrast, the \textbf{SCURank} model captures a unique semantic unit: the irony of their meeting origin (``Bennett contacted him to correct something he had written about her'').
This specific detail, while not containing high-frequency controversial terms, adds significant narrative completeness to the summary.
This observation supports our quantitative findings that SCURank, by leveraging Summary Content Units (SCUs), effectively identifies and retains distinct semantic information that might be overlooked by ranking methods relying solely on surface-level overlap or LLM-based preference, which can be biased towards sensational content.

\begin{table*}
  \centering
  \begin{tabular}{|p{\linewidth}|}
    \toprule
    \textbf{Article} \\
    Historical Chinese documents have helped scientists to track the decline of the world's rarest primates.  
    Today, China has between 26 and 28 Hainan gibbons left, but government records that date back to the 17th Century show that gibbons were once widespread across half of the country.  
    The apes began to disappear from the documents about 150 years ago, corresponding with population growth.  
    The study is published in the Proceedings of the Royal Society B.  Hainan gibbons are now limited to a few isolated patches of forest in the south west of China.  
    They live in just four social groups, one of which was only discovered a few weeks ago.  
    ...
    Dr Sam Turvey, from the Zoological Society of London, said: "China is one of the few places in the world that has a very very rich, long historical record. 
    ...
    These included records of animals, including gibbons, he said. 
    "We looked at the pattern of disappearance of gibbons through time and how that varied from place to place and the different environmental conditions and human pressures that were also present in these places."  
    The archives show that gibbons were a common sight in about 20 provinces in China well into the 17th and 18th Century.  
    However, Dr Turvey said it was "a stark contrast to their very imperilled position today".  
    "We see a really steep increase in population decline and real population collapse across China about 100-150 years ago," he added.  
    ...
    The researchers said a better understanding of the animals' decline would help them to establish a conservation plan for the country's last few Hainan gibbons.  
    Dr Turvey said: "It is an incredible privilege to be able to see gibbons in China in the wild.  
    "The Hainan gibbon is such as rare species, but knowing that this species is still hanging on there gives you hope that conservation will be able to bring that population back from the brink."  Follow Rebecca on Twitter    \\\midrule
    \textbf{SCURank distilled model's Output} \\
    Researchers have used historical Chinese documents to identify the decline of Hainan gibbons, which are now limited to isolated areas in China, amid population growth and environmental pressures.
    \\\midrule
    \textbf{MLE distilled model's Output} \\
    Historical Chinese documents have helped scientists understand the decline of the Hainan gibbon, revealing that gibbons were once widespread across half of the country, but are now limited to a few isolated patches.
    \\\midrule
    \textbf{Human-written Summary 1} \\
    Thanks to China's extensive historical records, scientists have been able to track the population decline of the rarest primate in the world. The Hainan gibbon was once widespread across the country, but due to deforestation and human population growth, there are now between 26 to 28 of the animals left.    \\
    \textbf{ROUGE-1: SCURank: 0.405 | MLE: 0.452} \\
    \midrule
    \textbf{Human-written Summary 2} \\
    Hainan gibbons in China have been decreasing precipitously in population ever since the 1800s, new records show. The decline accelerated in the 20th century with massive human population growth and deforestation. Today China has roughly 28 of the gibbons left.    \\
    \textbf{ROUGE-1: SCURank: 0.324 | MLE: 0.247} \\
    \midrule
    \textbf{Human-written Summary 3} \\
    Historical Chinese government documents show that although the world's rarest primates were once widespread across half of the country, their numbers are now declining.  The remaining primates, Hainan Gibbons, are now isolated to small patches of forest in China's southwest region.  Researchers continue to study the decline in the species. \\
    \textbf{ROUGE-1: SCURank: 0.400 | MLE: 0.541} \\
    \bottomrule
  \end{tabular}
  \caption{
    Sample summaries generated by the SCURank- and MLE-distilled models, together with the corresponding human-written summaries.
    ROUGE-1 scores for the SCURank and MLE models are also reported.
  }
  \label{tab:sample-summaries}
\end{table*}

\begin{table*}
  \small
  \centering
  \begin{tabular}{p{0.95\linewidth}}
  \toprule
  \textbf{Article} \\
  Green party leader Natalie Bennett has been forced to distance herself from a bizarre online site run by her boyfriend. Jim Jepps used a blog called The Daily Maybe to defend `rape fantasies', describe paedophiles as `complex human beings' and question why teachers who have relationships with pupils are put on the sex offenders register. \textbf{The couple met five years ago when Ms Bennett contacted him to correct something he had written about her}, but now the Green party stresses they do not `want to be associated' with his internet rants. ... (Context omitted for brevity) ... Ms Bennett has been dating Mr Jepps for five years, after meeting online. ... \\
  \midrule
  \textbf{SCURank Distilled Model} \\
  Natalie Bennett, the leader of the UK's Green party, has distanced herself from a website run by her boyfriend, Jim Jepps, who wrote controversial blog posts about sex, rape, and paedophiles. The couple met five years ago and \textbf{Bennett contacted him to correct something he had written about her}. The Green party has said that they do not want to be associated with his internet rants. \\
  \midrule
  \textbf{GPTRank Distilled Model} \\
  Natalie Bennett, the leader of the Green party, has distanced herself from a blog run by her boyfriend, Jim Jepps, who used the site to defend `rape fantasies' and describe paedophiles as `complex human beings'. The Green party has stressed that they do not want to be associated with his internet rants. Bennett has been dating Jepps for five years, but little is known about her private life. \\
  \bottomrule
  \end{tabular}
  \caption{Qualitative comparison of summaries generated by SCURank and GPTRank. \textbf{Bold text} highlights a specific Semantic Content Unit (SCU) regarding the origin of the couple's relationship. SCURank successfully captures this unique semantic detail, whereas GPTRank focuses primarily on high-frequency controversial keywords.}
  \label{tab:case_study_scurank_vs_gptrank}
  \end{table*}

  \section{LLMs Result}
  \label{sec:llms-result}
  \subsection{LLMs generated summaries}
  We used the same LLMs as in the LLMs-9 dataset to generate summaries for the test dataset and evaluate their performance.
  Table~\ref{tab:llms-result} presents the performance of various LLMs on both of the CNN/DailyMail and XSum datasets using multiple evaluation metrics and extractiveness measures.
  
  \subsection{LLM-based Summary Comparison Prompt}
  The following prompt was used to instruct each LLM to compare two summaries based on their information richness and importance.
  
  \begin{tcolorbox}[colback=gray!10, colframe=gray!50, boxrule=0.8pt, width=\columnwidth]
  \small
  Compare the following summaries based on **information richness** (how much relevant detail is preserved) and **importance** (focus on the most significant points).\\
  \\
  For each summary, evaluate:\\
  - Information richness: Comprehensiveness, specific details, coverage of key topics\\
  - Importance: Focus on high-impact information, critical insights, actionable content\\
  \\
  Provide a brief analysis and rank the summaries from best to worst, explaining your reasoning.\\
  \\
  Input format:\\
  Article: [article text]\\
  Summary 1: [summary 1 text]\\
  Summary 2: [summary 2 text]\\
  
  Output format:\\
  Summary 1 Analysis: [Your analysis here]\\
  Summary 2 Analysis: [Your analysis here]\\
  Winner: [1 or 2 or "tie" if they are equally good]\\
  \end{tcolorbox}  

\begin{table*}
  \centering
  \begin{tabular}{l|cccccc|cc}
    \toprule
    & R-1 & R-2 & R-L & BLEURT & BS & BaS & Coverage & Density\\
    \midrule\midrule
    \multicolumn{9}{c}{CNN/DailyMail}\\
    \midrule
    GPT-4o-mini       & 38.3 & 15.7 & 24.8 & 54.6 & 68.7 & -2.71 & 0.81 & 2.32\\
    GPT-4o            & 41.2 & 15.9 & 26.0 & 54.7 & 69.4 & -2.78 & 0.81 & 2.47\\
    GPT-4-turbo       & 41.2 & 15.9 & 26.0 & 54.7 & 69.4 & -2.78 & 0.81 & 2.47\\
    Gemini-1.5-flash  & 43.1 & 17.9 & 28.0 & 55.5 & 70.4 & -2.40 & 0.83 & 2.53\\
    Gemini-1.5-pro    & 41.5 & 16.7 & 27.2 & 53.9 & 69.7 & -2.67 & 0.79 & 2.26\\
    Llama-3.1-70b     & 42.7 & 18.4 & 28.0 & 54.4 & 69.7 & -2.47 & 0.86 & 3.24\\
    Llama-3.1-402b    & 42.7 & 19.0 & 28.1 & 55.2 & 69.9 & -2.39 & 0.86 & 3.25\\
    Mistral-Large     & 45.1 & 18.8 & 29.4 & 55.9 & 71.2 & -2.48 & 0.85 & 2.77\\
    Claude-3.5-sonnet & 42.3 & 16.9 & 27.2 & 51.5 & 69.5 & -2.65 & 0.83 & 3.01\\
    \midrule\midrule
    \multicolumn{9}{c}{XSum}\\
    \midrule
    GPT-4o-mini       & 37.1 & 14.3 & 24.9 & 53.2 & 67.7 & -2.87 & 0.80 & 2.53\\
    GPT-4o            & 38.6 & 14.4 & 24.8 & 50.5 & 67.6 & -2.93 & 0.78 & 2.38\\
    GPT-4-turbo       & 38.6 & 14.4 & 24.8 & 50.5 & 67.6 & -2.93 & 0.78 & 2.38\\
    Gemini-1.5-flash  & 42.7 & 18.2 & 29.4 & 55.0 & 69.9 & -2.60 & 0.81 & 2.73\\
    Gemini-1.5-pro    & 41.3 & 15.3 & 27.7 & 53.8 & 69.1 & -2.75 & 0.78 & 2.24\\
    Llama-3.1-70b     & 40.4 & 14.7 & 25.9 & 53.1 & 68.7 & -2.63 & 0.83 & 2.97\\
    Llama-3.1-402b    & 42.4 & 17.8 & 28.8 & 54.8 & 69.6 & -2.51 & 0.86 & 3.42\\
    Mistral-Large     & 44.4 & 17.7 & 29.8 & 54.5 & 69.8 & -2.64 & 0.83 & 2.89\\
    Claude-3.5-sonnet & 42.5 & 17.1 & 28.1 & 53.2 & 69.6 & -2.69 & 0.83 & 2.97\\
    \bottomrule
  \end{tabular}
  \caption{
    Automatic evaluation results of summaries generated by various LLMs on CNN/DailyMail and XSum datasets.
    Evaluation metrics include R-1(ROUGE-1), R-2(ROUGE-2), R-L(ROUGE-L), BLEURT, BS(BERTScore), BaS(BartScore), Coverage, and Density.
  }
  \label{tab:llms-result}
\end{table*}

\end{document}